\pdfoutput=1

\documentclass[11pt]{article}

\usepackage{acl}

\usepackage{multirow}
\usepackage{times}
\usepackage{latexsym}
\usepackage{booktabs}
\usepackage{graphicx}
\usepackage{subcaption}
\usepackage{xcolor}
\usepackage{soul}
\usepackage{varioref}
\usepackage{amsmath}
\usepackage{cleveref}
\usepackage{times}
\usepackage{tensor} 
\usepackage{latexsym}
\usepackage{multirow}
\usepackage{booktabs}
\usepackage{amssymb}
\usepackage{graphicx}
\usepackage{chemformula}

\usepackage[utf8]{inputenc}
\usepackage{enumitem}
\crefname{section}{\S}{\S\S}
\Crefname{section}{\S}{\S\S}

\usepackage[T1]{fontenc}

\usepackage[utf8]{inputenc}

\usepackage{microtype}
\def\LatinUpper{A,B,C,D,E,F,G,H,I,J,K,L,M,N,O,P,Q,R,S,T,U,V,W,X,Y,Z}
\def\LatinLower{a,b,c,d,e,f,g,h,i,j,k,l,m,n,o,p,q,r,s,t,u,v,w,x,y,z}
\makeatletter
\newcommand{\genCal}[1]{\expandafter\newcommand\csname c#1\endcsname{{\mathcal #1}}}
\@for\i:=\LatinUpper\do{%
	\expandafter\genCal\i
}
\renewcommand{\vec}[1]{{\mathbf{#1}}}
\newcommand{\genLatinVec}[1]{\expandafter\newcommand\csname v#1\endcsname{{\vec #1}}}
\@for\i:=\LatinLower\do{%
	\expandafter\genLatinVec\i
}
\usepackage{caption}
\newcommand{\bc}[1]{\left\{{#1}\right\}}
\newcommand{\br}[1]{\left({#1}\right)}
\newcommand{\bs}[1]{\left[{#1}\right]}
\title{Shapes of Emotions: Multimodal Emotion Recognition in Conversations via Emotion Shifts}

\author{Harsh Agarwal\thanks{\ \ Equal Contributions} \qquad Keshav Bansal\footnotemark[1] \qquad \textbf{Abhinav Joshi} \qquad \textbf{Ashutosh Modi} \\
Indian Institute of Technology Kanpur (IIT-K) \\ 
 \texttt{\{harshagarwal0194,keshav22bansal\}@gmail.com} \\
\texttt{\{ajoshi,ashutoshm\}@cse.iitk.ac.in}
}

\begin{document}
\maketitle
\begin{abstract}
Emotion Recognition in Conversations (ERC) is an important and active research area. Recent work has shown the benefits of using multiple modalities (e.g., text, audio, and video) for the ERC task. In a conversation, participants tend to maintain a particular emotional state unless some stimuli evokes a change. There is a continuous ebb and flow of emotions in a conversation. Inspired by this observation, we propose a multimodal ERC model and augment it with an emotion-shift component that improves performance. The proposed emotion-shift component is modular and can be added to any existing multimodal ERC model (with a few modifications). We experiment with different variants of the model, and results show that the inclusion of emotion shift signal helps the model to outperform existing models for ERC on MOSEI and IEMOCAP datasets. 
\end{abstract}

\section{Introduction}
\noindent Humans are complex social beings, and emotions are indicative of not just their inner state and feelings but also their internal thinking process \cite{minsky2007emotion}. To fully understand a person, one needs to understand their inherent emotions. Recent research has witnessed colossal interest in including artificially intelligent machines as conversable companions for humans, e.g., personal digital assistants. However, communication with AI systems is quite limited. AI systems do not understand the inherent emotions expressed implicitly by humans making them unable to comprehend the underlying thought processes and respond appropriately. Consequently, a wide variety of approaches have been proposed for developing emotion understanding and generation systems \cite{sharma2021survey, witon2018disney, singh-etal-2021-end, goswamy2020adapting, colombo2019affect, singh2021fine, joshi-etal-2022-cogmen}. 

\begin{figure}[t]
\centering
\includegraphics[scale=0.36]{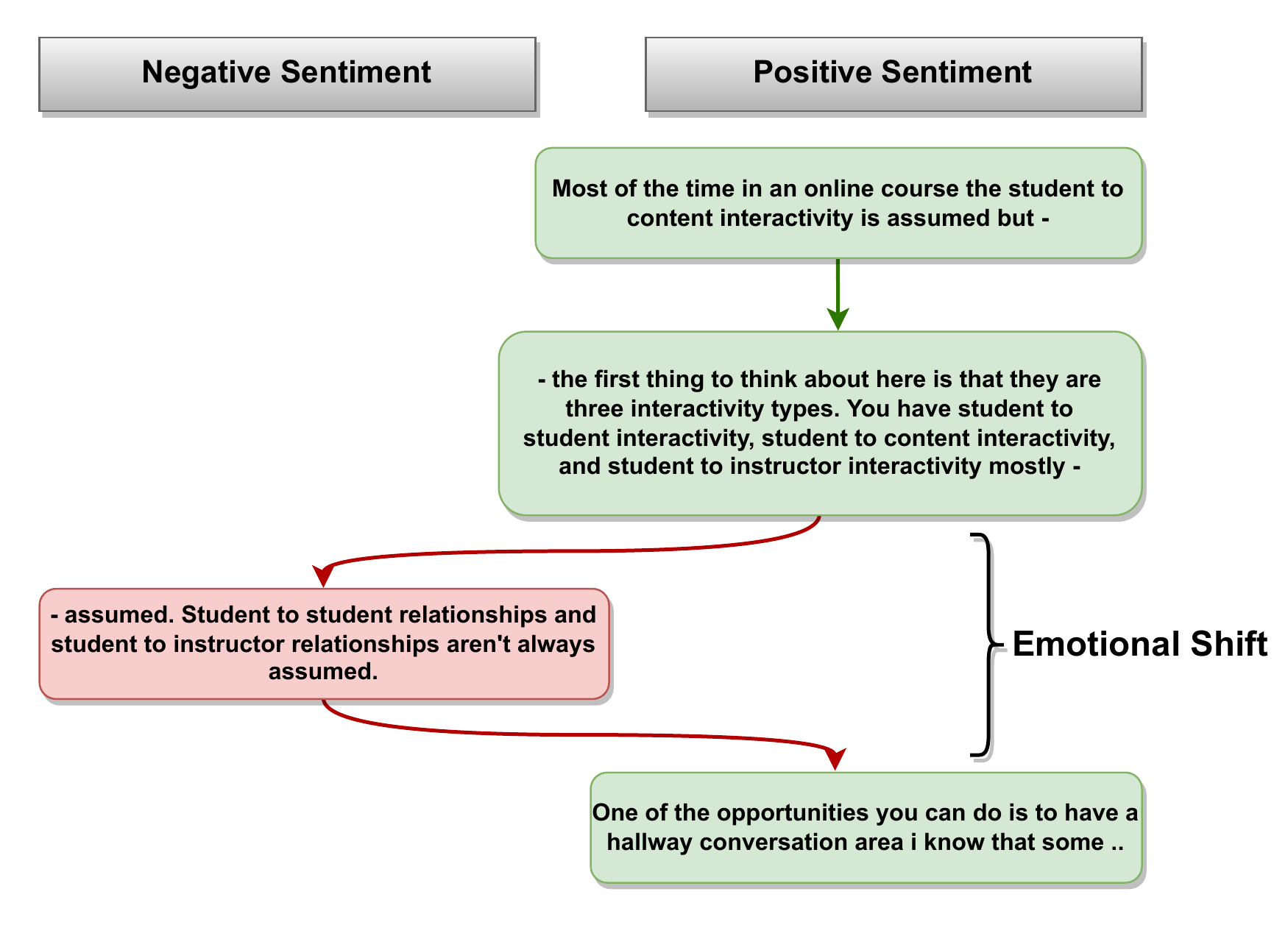}
\caption{Emotional shift on the dialogue ``m7SJs73SF8w'' from CMU-MOSEI dataset}
\label{fig:emotional_shift_example}
\vspace{-6mm}
\end{figure}
During an interaction, humans express different emotions and fluctuate between multiple emotional states. It is often the case that participants in a conversation tend to maintain a particular emotional state unless some stimuli evokes a change. This observation is closely related to \textit{Shapes of Stories} proposed by renowned writer Kurt Vonnegut \cite{vonnegut1995shapes}, who posits that every story has a shape plotted by the ups and downs experienced by the characters of the story, and this, in turn, defines an \textit{Emotional Arc} of a story. This phenomenon has also been empirically verified by \citet{reagan2016emotional}, who analyzed around 1300 stories to come up with common emotional arc patterns across various stories. Moreover, apart from these flows, there exists a sudden shift of emotions from positive to negative sentiments. Consider an example shown in Figure \ref{fig:emotional_shift_example}, where the sentiment of the third utterance shifts from positive to negative and back again to positive in the fourth utterance. 
Current state-of-the-art methods are often oblivious to the presence of such emotion shifts and tend to fail in cases where there is a sudden change in the emotional state  \cite{survey_paper}. To address this issue, we propose incorporating a novel module that explicitly tracks such emotional shifts in conversations.
Humans express their emotions via various modalities, such as language, modulations in voice, facial expressions, and body gestures. 
In this paper, to fully and correctly recognize human emotions, we propose a multimodal emotion recognition system that utilizes language, audio, and video modalities. 
We propose a multimodal ERC model based on GRUs that fuses information from different modalities. An independent emotional shift component 
captures the emotion shift signal between consecutive utterances, 
allowing the model to forget past information in case of an emotional shift. 
We make the following contributions: 
\begin{itemize}[noitemsep,nolistsep]
\item We propose a new deep learning based multimodal emotion recognition model that captures information from text, audio, and video modalities. We release the model implementation and experiments code:  \url{https://github.com/Exploration-Lab/Shapes-of-Emotion} 
\item We propose a novel emotion shift network (modeled via a Siamese network) that guides the main emotion recognition system by providing information about possible emotion shifts or transitions. The proposed component is modular, it can be pretrained and added to any existing multi-modal ERC (with a few modifications) to improve emotion prediction. 
\item The proposed model is experimented on the two widely known multimodal emotion recognition datasets (MOSEI and IEMOCAP), and results show that emotional shift component helps to outperform some of the existing models. We perform detailed analysis and ablation studies of the model and show the contribution of different components. We analyse the performance of our model in the classification of utterances having a shift in emotion and compare this with previous models and report an improvement due to the use of emotion-shift information. We further examine how the internal GRU gates behave during emotion shifts. 
\end{itemize}
\section{Related Work} \label{sec:relatedwork}

\begin{figure*}[t]
\centering
\includegraphics[scale=0.30]{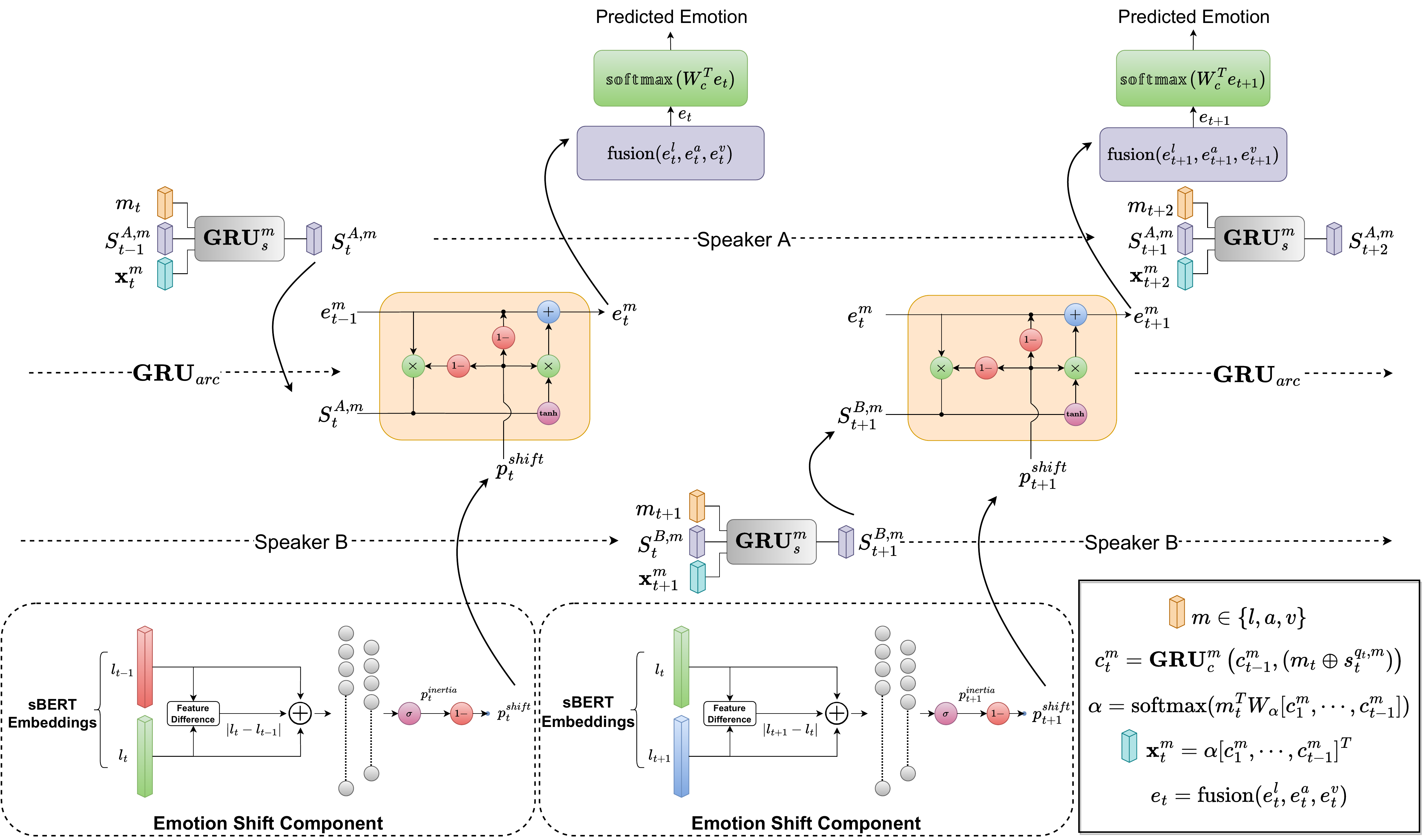}
\caption{The model architecture for a conversation between two speakers, A and B, at time $t$ and $t+1$. The upper part highlights the Emotion Classification Component, and the lower part highlights the Emotion Shift Component.}
\label{fig:model}
\end{figure*}

Emotion recognition using multiple modalities is an active area of research leading to the development of widely popular benchmark datasets, e.g., CMU-MOSEI \cite{mosei}, and IEMOCAP \cite{iemocap}. Recent works have highlighted the crucial aspects of self, and interpersonal dependencies in the emotional dynamics of the conversations \cite{survey_paper}. Another essential feature is the role of the local and global context for emotion recognition systems. Some notable works like Dialogue RNN \cite{dialogue_rnn} try to capture these properties by modeling each speaker with a party state and the emotion of each utterance by an emotional state. Furthermore, a context state is maintained to model the global conversation context. Another work Multilogue-Net \cite{multilogue_net} highlights the limitation of the fusion mechanism used in Dialogue RNN \cite{dialogue_rnn} and tries to solve it using a party, context, and emotion GRUs for each modality. It uses a pairwise attention mechanism proposed by \cite{pairwise_attention} to fuse the emotion states for all the modalities effectively. However, DialogueRNN highlights the poor performance in predicting the emotions with the utterances where the emotion shifts from positive to negative sentiments. Our work considers the emotion shifts present in the dialogues and tries to leverage them for improving emotion recognition. Another line of work includes Transformer-Based Joint-Encoding (TBJE) \cite{mosei_umons} that achieves the state-of-the-art results on the sentiment task for the MOSEI dataset using a multimodal transformer-based model for combining multiple modalities. However, in the emotion task, TBJE is outperformed by the Multilogue-Net model. The possible reason highlighted by the paper is the lack of context-awareness in the architecture, as TBJE neither uses the previous nor next utterance to predict the emotion for the current utterance. Some of the other works in multimodal emotion recognition include the Memory Fusion network (MFN) \cite{zadeh2018memory_MFN}, which aligns multimodal sequences using multi-view gated memory, Graph-MFN \cite{mosei} which uses Dynamic Fusion Graph (DFG) and learns to model the n-modal interactions dynamically, and bc-LSTM \cite{poria2017} which uses an LSTM-based model to capture contextual information. CESTa \cite{wang-etal-2020-contextualized} captures the emotional consistency in the utterances using CRF model \cite{10.5555/645530.655813ConditionalRandomFields} for boosting the performance of emotion classification and comes close to our idea of leveraging emotion shifts. 


\section{Task and Corpus}
\label{sec:methadology}
\noindent \textbf{Problem Definition:} Consider a conversation having utterances $u_1, \ldots, u_{N}$. 
The task of Emotion Recognition in Conversation (ERC) is to predict the emotion (or sentiment) of each utterance $u_{t}$. We define an utterance to be a coherent piece of information (single or multiple sentences) conveyed by a single participant at a given time. 
We model an utterance in terms of different modalities: $u_{t} = \{l_t,a_t,v_t\}$. An utterance ($u_{t}$) at time-step $t$ is represented via features 
from textual transcript ($l_{t}$), audio ($a_{t}$), and visuals ($v_{t}$) of the speaker. We denote the speaker of utterance $u_t$ as $q_t$. 
\subsection{Corpus Details}
\label{sec:corpus}
\subsubsection{CMU-MOSEI:} The CMU Multimodal Opinion Sentiment and Emotion Intensity \cite{mosei} is an English language dataset containing more than 65 hours of annotated video from more than 10000 speakers and 250 topics. Each sentence is annotated for a sentiment on a $[-3,3]$ Likert scale. However, in this work, we project these labels to a two-class classification setup with values $\ge$ 0 signifies positive sentiments and values $<$ 0 convey negative sentiments. Dataset also contains six emotion labels, namely angry, happy, sad, surprise, fear and disgust for each utterance. Note that in case of emotions labels the utterances are multi-label. Which means a single utterance can have more than one emotion label.
We have shown results for both sentiment and emotion prediction tasks. 
\subsubsection{IEMOCAP:} The IEMOCAP benchmark \cite{iemocap} consists of a conversation between ten distinct speakers. The dataset contains two-way conversations in videos where every video clip contains a single dyadic English dialogue. Further, each dialogue segments into utterances with an emotion label from six emotion labels, i.e., happy, sad, neutral, angry, excited, and frustrated. The dataset incorporates an acted setting where actors perform improvisations or scripted scenarios, specifically selected to elicit emotional expression. 
\section{The Proposed Model} \label{subsec:proposed_model}
\noindent During a conversation, speakers tend to maintain an emotional flow of affective states. These states majorly rely on the context of the entire conversation; for example, if the overall gist of the speaker about the topic is positive, the emotions like happiness, joy, and surprise can be seen more often than negative emotions like anger and sadness. Moreover, a speaker's emotions are often affected by the past emotions present in the conversation. Hence, an emotion prediction model should not only take into account the context but should also be able to maintain the speaker-level information along with the emotions present in the past utterances. Considering these assumptions, we propose the primary component of our emotion recognition model: the \textit{Emotion Classification Component}. The emotions classification component predicts an utterance's emotion label using information from the current speaker, emotions of the previous utterances, and the entire conversation context.
Another significant insight about the emotions in a conversation is the sudden shift of emotional states. Many of the existing state-of-the-art approaches highlight this in their error analysis, where the model fails to capture the sudden shifts in emotional states leading to a misclassified emotion prediction. To incorporate the effect of a sudden shift in the emotion, we introduce a separately trained component called the \textit{Emotion Shift Component}. The emotion shift component explicitly models the probability ($p^{shift}_t$) of a shift in emotion between the utterances $u_{t-1}$ and $u_{t}$. This shift in emotion can be expressed as moving from a positive (e.g., happy) to negative emotion (e.g., sad) or vice-versa between consecutive utterances. The emotion shift component being independent of the primary architecture is pretrained, and helps control the information flow from past to future during a sudden change. The signal from the pretrained emotional shift component is added to the emotion classification component to control the flow of emotions from past to future. Figure \ref{fig:model} shows detailed architecture of the proposed model.  
\noindent\textbf{Emotion Classification Component:}
\noindent For modeling the underlying emotions in a conversation, we maintain a \textit{party state}, \textit{emotion state} and \textit{context state}. The party state is maintained for each speaker and helps to keep track of the participant specific aspect in a conversation. The context state is global (common across each participant) and helps to encode the entire conversation context, thereby capturing inter-utterance dependencies. Akin to the context state, the emotion state is also global and helps to leverage the emotion information flow between utterances. Moreover, the emotion shift signal between the current and previous utterance is used to update the global emotion state. The emotion label for each utterance is then predicted by decoding the emotion state. In our model each of the party, context and emotion states are modality specific and are updated using a modality specific GRU \cite{chung2014empirical} network for each modality $m\in\{l,a,v\}$ (indicated by the superscript $m$). We employ late fusion to combine the emotion states from different modalities. 
Next, we explain different GRU networks used in the model. 
\begin{align}
s^{q_t,m}_{t} &= \textbf{\text{GRU}}_s^{m}\left(s^{q_{t},m}_{t-1},\left(m_{t} \oplus \mathbf{x}_{t}^{m}\right)\right) \label{eqn:sgru}\\
c_{t}^{m} &= \textbf{\text{GRU}}_c^{m}\left(c_{t-1}^{m},\left(m_{t} \oplus s_{t}^{q_t, m}\right)\right) \label{eqn:cgru} \\
e_{t}^{m} &= \textbf{\text{GRU}}_{arc}^{m}\left(e_{t-1}^{m}, s_{t}^{q_t,m}, p_t^{\text{shift}}\right) \label{eqn:egru_base} \\
e_t &= \text{fusion}(e^l_t, e^a_t, e^v_t) \label{eqn:efusion}
\end{align}
\noindent \textit{Party State Update (GRU$_s$):} The state of each participant is modeled by the party state update GRU$_s$. 
For each modality $m\in\bc{l,a,v}$, $q_t$'s party state $s^{q_t,m}_{t-1}$ is updated to $s^{q_t,m}_{t}$ using an attention vector $\mathbf{x}^m_t$  and modality specific feature $m_{t}$ (Eq. \ref{eqn:sgru}), $\oplus$ denotes concatenation operation.  Here $\mathbf{x}^m_t$ is calculated using a simple dot product attention mechanism over the context states ($c^m_{t}$). Note that for all speakers other than $q_t$, the party state at $t-1$ and $t$ remains the same. 

\noindent\textit{Context State Update (GRU$_c$):} Global conversation context is modeled using the context state update GRU$_c$. For each modality $m\in\bc{l,a,v}$, the global context state $c^m_{t-1}$ is updated to $c^m_{t}$ (Eq. \ref{eqn:cgru}) using the $q_t$'s party state $s^{q_t,m}_t$ and the corresponding modality feature $m_t$. 
Context states ($c_1^m,\cdots,c_{t-1}^m$) are used for calculating the attention vector $x_t^m$ for each modality $m\in\bc{l,a,v}$ as follows: 
\begin{align}
\alpha &= \operatorname{softmax}\br{m_t^TW_\alpha\bs{c_1^m,\cdots, c_{t-1}^m}} \label{eqn:alpha}\\
x_t^m &= \alpha\bs{c_1^m,\cdots, c_{t-1}^m}^T\label{eqn:alpha2}
\end{align}



\noindent \textit{Emotion State Update (GRU$_{arc}$):} For each modality $m \in \bc{l,a,v}$, the global emotion state $e^m_{t-1}$ is updated to $e^m_{t}$ (Eq. \ref{eqn:egru_base}) using the current party state $s^{q_t,m}_t$ and modulated by the emotion shift component ($p_t^{\text{shift}}$). 
%
%
%
%
%
%
\noindent The emotion states for all the three modalities are fused together (Eq. \ref{eqn:efusion}) to create $e_t$ using a pairwise attention mechanism \cite{multilogue_net}. $e_t$ is later used to decode the emotion class for an utterance.

The emotion classification component is a context-aware model similar to that of previous works like Multilogue-net \cite{multilogue_net} but with a few key differences. Firstly, instead of modelling an emotion state for each participant, we introduce global emotion state for each conversation. This is done to make use of the flow of emotion between utterances. Secondly, the emotion shift signal between the current and previous utterance ($p_t^{\text{shift}}$) is used to update the global emotion state using a GRU$_{arc}$ which aims to model the emotion arc in the conversation.

\noindent\textbf{Emotion Shift Component:}
\noindent To capture the emotional arc across the conversation, we explicitly model probability of emotion shift ($p_{t}^{shift}$) between successive utterances ($u_{t-1}$ and $u_{t}$). We use a Siamese network \cite{siamese_main} to model the emotional shift present across utterances. A Siamese network generally consists of two or more identical subnetworks having the same configuration with shared parameters and weights. The proposed emotion shift architecture takes the textual features of the current ($l_{t}$) and previous ($l_{t-1}$) utterances and outputs the probability of maintaining emotional inertia ($p_t^{inertia}$). The architecture of the emotion shift prediction network is shown in lower half of Figure \ref{fig:model}. We use Sentence-BERT (SBERT) \cite{sbert} embeddings as textual features. SBERT is a modification of the pre-trained BERT \cite{bert} network that uses the Siamese network to derive semantically meaningful sentence embeddings for transfer learning tasks. The emotion shift prediction network makes use of only the text modality for two reasons. Firstly, it has often been found empirically that among the text, audio, and video modalities, text modality carries more information for ERC tasks  \cite{mutlimodal_survey_poria_iemocap}. Secondly, early fusion techniques to combine the three modalities can suffer in a Siamese-type architecture due to difficulty in mapping the fused modality vector to a vector space in which similar vectors are close. We also experimentally verify this (\S\ref{sec:ablation_study}). 

The emotion shift prediction network (between $u_{t}$ and $u_{t-1}$)  takes in text features corresponding to utterances  ($l_{t}$ and $l_{t-1}$) and their element wise differences to output the  probability of a shift as given by (Eq. \ref{eqn:emotion_shift_prob}). Here, $p_t^{inertia}$ is calculated using Siamese network (Eq. \ref{eqn:emotion_interia_prob}, \ref{eqn:siamese}). Here, $\cH_{t}$ is the Siamese hidden state, $\cW$ $\left(\in \mathbb{R}^{3d_l}\right)$ is the model parameter. For the Siamese network, we use Binary Cross Entropy loss ($\cL_{{s}}$) over the distribution $p_t^{\text{shift}}$. The emotion shift component modulates the Emotion State GRU$_{arc}$ via $p_t^{\text{shift}}$ and hence controls the flow of information during the conversation. The Emotion Shift component captures the emotional consistency in the utterances and can act as an independent modular component that can be pretrained and added to any existing multi-modal ERC framework with a few modifications for improving emotion recognition in conversations.
\begin{align}
p_t^{\text{shift}} &= 1 - p_t^{inertia} \label{eqn:emotion_shift_prob} \\
p_t^{inertia} &= \sigma\br{\cH_t} \label{eqn:emotion_interia_prob} \\
\cH_{t} &= {\cW}^T(l_{t-1} \oplus l_t \oplus \left|l_t - l_{t-1}\right|) \label{eqn:siamese}
\end{align}
\noindent \textbf{Overall Architecture:}
The motivation for the proposed architecture follows from the intuition that we need to weigh down the contribution of the previous emotion state in case of an emotion shift. In other words, we need to reduce the influence of $e^m_{t-1}$ in the calculation of $e^m_t$ when there is a high $p^{shift}_t$. To do so, we modify the reset and update gates in the GRU modelling the emotional arc of the conversation i.e. GRU$_{arc}$. A GRU has gating units (reset and update gates) that modulate the flow of information inside the unit. \citet{speech_recognition_gru} mention the usefulness of reset gate in scenarios where significant discontinuities are present in the sequence, thereby indicating its crucial role to forget information. Their work also finds a redundancy in the activations of the reset and update gates when processing speech sequences. Motivated by this, and the intuition that we need to forget more information when there is a higher probability of an emotional shift, we directly use the value of $(1-p^{shift}_t)$ for both the reset and update gates. The updates for GRU$_{arc}$ unit are given by Eqs. \ref{eqn:gru_candidate}, \ref{eqn:gru_interpolate}. Eq. \ref{eqn:gru_candidate} calculates a candidate emotion state $\tilde{e}^m_t$ in which the prior emotion state's ($e^m_{t-1}$) is controlled by the emotion shift signal. The output $e^m_t$ is a linear interpolation between $\tilde{e}^m_{t-1}$ and $e^m_{t-1}$. Again, $p_t^{shift}$ controls the influence of $e^m_{t-1}$ (Eq. \ref{eqn:gru_interpolate}). Therefore, a higher value of $p^{shift}_t$ will limit the contribution of the previous emotion state.
In the absence of the emotion shift component, the GRU gates are learned using only the classification data, much like the rest of the parameters in the model. If the total number of parameters in a model is huge (as is the case with most deep learning models), the gates might be unable to learn well. We verify that the modeling of  the shift in emotion encourages better learning of these gates (\S\ref{sec:ablation_study}).
\begin{align}
	\tilde{e}^m_t &= \operatorname{tanh}\br{W {}s^{q_t,m}_t  + (1-p^{shift}_t) \odot \br{Ue^m_{t-1}}} \label{eqn:gru_candidate}\\
	e^m_t &= (1-p^{shift}_t) \odot e^m_{t-1} + p^{shift}_t \odot \tilde{e}^m_t \label{eqn:gru_interpolate} 
\end{align}

For prediction at time $t$, the emotion vector $e_t$ (formed from fusion of $e_t^{m}$ as described in (Eq. \ref{eqn:efusion})) is passed through a final classification layer $W_c$ ($\in \mathbb{R}^{d_e\times K}$) where $K$ is the number of emotion or sentiment classes. This is used to obtain probability distribution over emotion labels via the Softmax activation: $o = \operatorname{softmax}(W_c^Te_t)$. We use the Cross-Entropy Loss over this distribution to train the weights. 

\begin{table}[t]
\centering
\small

\begin{tabular}{ccccc} 
\toprule
\multirow{2}{*}{Dataset} & \multicolumn{2}{c}{\#utterances} & \multicolumn{2}{c}{Emotion shift (in \%)}  \\ 
\cmidrule{2-5}
                         & Train & Test                     & Train & Test                       \\ 
\midrule
CMU-MOSEI                & 18191 & 4655                     & 33.61 & 34.62                      \\
IEMOCAP                  & 5810  & 1623                     & 12.89 & 12.75                      \\
\bottomrule
\end{tabular}

\caption{Statistics for number of utterances and emotion shift percentage in various datasets}
\label{table:emotion_shift_data}
\end{table}

\section{Experiments and Results}

\noindent\textbf{Multimodal Emotion Corpora:} We evaluate our model using two benchmark English  ERC  datasets - CMU Multimodal Opinion Sentiment and Emotion Intensity (CMU-MOSEI) dataset and the Interactive Emotional Dyadic Motion Capture (IEMOCAP) dataset. 
Details of these corpora are discussed in \S \ref{sec:corpus}.
In a nutshell, each of the two corpora has language, audio, and video modalities. MOSEI has both sentiment and six emotion labels, IEMOCAP has video recordings of dyadic conversations and is labeled with six emotion labels.

\noindent\textbf{Emotion shift in Dataset:}
We define an emotion shift between consecutive utterances if there is a shift from a positive to a negative emotion or vice-versa. CMU-MOSEI dataset provides annotated (positive/negative) sentiment label for each utterance. This is not the case for the IEMOCAP dataset, therefore we divide the emotion classes into a positive and negative category. Happiness and surprise are taken into the positive category while disgust, angry and sad are considered as the negative category. Note that IEMOCAP also has a neutral emotion, but a shift is only counted if it is from a positive to negative emotion or vice-versa. Table \ref{table:emotion_shift_data} shows the percentage of emotion shift observed in the datasets. Since CMU-MOSEI shows a larger amount of emotion shift, we were motivated to perform experiments on CMU-MOSEI first.

\subsection{Results}

\begin{table}[t]
\centering
\small
\renewcommand{\arraystretch}{1}
\setlength\tabcolsep{4.1pt}
\begin{tabular}{lcc} 
\toprule
Model & F1 & Accuracy  \\ 
\midrule
Graph-MFN      & 77.00             & 76.90       \\
DialogueRNN    & 79.82             & 79.98       \\
Multilogue-Net & 80.01             & 82.10    \\
TBJE    & -              & 82.4           \\
Our Model      & \textbf{83.07}            &\textbf{82.66}       \\
\bottomrule
\end{tabular}
\caption{Performance comparison on the sentiment task of CMU-MOSEI dataset (all numbers in \%)}
\label{table:mosei_result}
\end{table}

\begin{table}[t]
\centering
\renewcommand{\arraystretch}{0.9}
\setlength\tabcolsep{4.1pt}
\small
\resizebox{\columnwidth}{!}{
\begin{tabular}{ccccccc} 
\toprule
\multirow{2}{*}{Emotion} & \multicolumn{2}{c}{Multilogue-Net} & \multicolumn{2}{c}{TBJE} & \multicolumn{2}{c}{Our Model}  \\ 
\cmidrule{2-7}
                         & A     & F1                         & A    & F1                       & A     & F1                     \\ 
\midrule
Happiness                & \textbf{70.05} & \textbf{70.03 }                     & 66.00   & 65.50                     & 68.51 & 68.61                  \\
Sadness                  & 71.04 & 70.42                      & 73.90 & 67.90                     & \textbf{74.20}  & \textbf{71.74}                  \\
Anger                    & 74.78 & 74.31                      & \textbf{81.90} & 76.00                       & 75.17 & \textbf{76.10 }                  \\
Disgust                  & 77.98 & 79.20                       & \textbf{86.50} & \textbf{84.50}                     & 83.67 & 82.79                  \\
Fear                     & 69.04 & 75.50                       & \textbf{89.20} & \textbf{87.20}                     & 87.11 & 85.90                   \\
Surprise                 & 88.89 & 85.98                      & \textbf{90.60} & \textbf{86.10}                     & 78.99 & 81.62                  \\
\bottomrule
\end{tabular}
}
\caption{Performance comparison on the emotion task of CMU-MOSEI dataset (all numbers in \%)}
\label{table:mosei_emotion_result}
\end{table}

\noindent We evaluate our approaches using standard F1 score and Accuracy evaluation metrics (App. \ref{appendix:metrics}). We train and report the performance of our model for four sub-tasks, 2-way sentiment classification and binary emotion classification on CMU-MOSEI, and four-class and six-class emotion classification task for IEMOCAP. The focus of our work is multimodal ERC and consequently, as is done in previous work, we compare only with previous multimodal approaches, since comparison with unimodal (e.g., text) only approaches does not make sense. Moreover, SOTA unimodal approaches (such as text based) use additional information such external knowledge sources (e.g., \cite{ghosal-2020-cosmic}) which makes the comparison with multimodal approach unfair specially given that such knowledge may not be available for other modalities. Nevertheless, it is possible to incorporate the emotion shift component into existing emotion prediction architectures (unimodal or multimodal) and we leave this exploration for future.  

\noindent\textbf{Results on CMU-MOSEI:} Table \ref{table:mosei_result} shows comparison of our best performing model on CMU-MOSEI sentiment labels, with current state of the art models: TBJE \cite{mosei_umons}, Multilogue-Net \cite{multilogue_net}, Dialogue RNN \cite{dialogue_rnn}, and Graph-MFN \cite{poria2017}. As evident from the results, we are able to significantly outperform the previous SOTA Multilogue-Net model with an increase of 3\% in F1 score.
We further compare our model on the emotion classification task with TBJE and Multilogue-Net (Table \ref{table:mosei_emotion_result}). As shown in the table, our model outperforms for some of the emotion classes. We speculate that poor performance is due to the multilabel setting in the CMU-MOSEI dataset. As the emotion labels are multilabel, the emotion shift component is not able to play a meaningful role in providing a performance boost to the emotion classification component. We consider multilabel settings as another line of future work where the emotion shift modeling takes into account the multilabel property.

\begin{table}[t]
\renewcommand{\arraystretch}{0.7}
\setlength\tabcolsep{5pt}
\centering
\small
\resizebox{\columnwidth}{!}{
\begin{tabular}{ccccccc} 
\toprule
\multirow{2}{*}{Emotion} & \multicolumn{2}{c}{bc-LSTM} & \multicolumn{2}{c}{CHFusion} & \multicolumn{2}{c}{Our model}  \\ 
\cmidrule{2-7}
                         & A     & F1                  & A    & F1                    & A     & F1                     \\ 
\midrule
Happy                    & \textbf{79.31} & -                   & 74.30 &\textbf{ 81.40}                  & 68.75 & 72.79                  \\
Sad                      & \textbf{78.30 }& -                   & 75.60 & 77.00                  & 76.73 & \textbf{81.21}                  \\
Neutral                  & 69.92 & -                   & 78.40 & 71.20                  & \textbf{81.51 }& \textbf{78.25  }                \\
Angry                    & 77.98 & -                   & 79.60 & 77.60                  &\textbf{ 82.35 }& \textbf{79.77}                  \\
Avg(w)                   & 75.20  & -                   & 76.50 & 76.80                  &\textbf{ 78.47} & \textbf{78.46}                  \\
\bottomrule
\end{tabular}
}

\caption{Performance comparison on the IEMOCAP dataset for four emotion labels (all numbers in \%)}
\label{result_iemocap_1}
\vspace{-3mm}
\end{table}

\begin{table}[t]
\renewcommand{\arraystretch}{0.7}
\setlength\tabcolsep{2.4pt}
\small
\begin{tabular}{cccccccc}
\toprule
Emotion  & Hap & Sad   & Neu & Ang & Exc & Fru & Avg(w) \\
\midrule
Acc & 54.17 & 65.31 & 62.50   & 62.94 & 67.89   & 64.04      & 63.59  \\
F1       & 50.81 & 70.48 & 60.23   & 63.69 & 70.73   & 62.72      & 63.82 \\
\bottomrule
\end{tabular}
\caption{Performance of our model on IEMOCAP dataset for 6 labels (all numbers in \%)}
\label{result_iemocap_2}
\vspace{-3mm}
\end{table}

\noindent\textbf{Results on IEMOCAP:} Previous works on Multimodal-IEMOCAP have shown performance only on angry, happy, sad, and neutral emotions. We compare our model performance on these four classes with state-of-the-art models CHFusion \cite{Hfusion} and bc-LSTM \cite{poria2017} (Table \ref{result_iemocap_1}). Our model significantly outperforms both of these on average weighted F1 and Accuracy. Also, emotion classes neutral and angry show improved performance. We also provide results on six emotion classes - happy, sad, neutral, angry, excited, and frustrated (Table \ref{result_iemocap_2}). For these experiments, we use BERT features for text (\S
\ref{sec:ablation_study}), OpenSmile features for audio and 3D-CNN features \cite{dialogue_rnn} for video. We did not come across any existing work on 6-class multimodal IEMOCAP for the comparison. 




\noindent\textbf{Performance of emotion shift component:} The results describing the capability of the emotion shift component to predict the shift for CMU-MOSEI and IEMOCAP dataset are shown in Table \ref{siamese_accuracy}. It is to be noted that predicting the shift accurately is not our primary objective. Our objective is to be able to improve the emotion prediction by using the signal ($p_t^{shift}$) received from the emotion shift component. 

\begin{table}[t]
\centering
\small
\renewcommand{\arraystretch}{0.7}
\setlength\tabcolsep{5pt}
\begin{tabular}{cccc} 
\toprule
\multicolumn{2}{c}{{Datasets}} &  Accuracy     & F1                         \\ 
\midrule
\multicolumn{2}{c}{CMU-MOSEI}                 & 72.65 & 67.32                      \\
\multirow{2}{*}{IEMOCAP} & 4-label            & 80.50 & 79.68                      \\
                         & 6-label            & 85.63 & 84.28                      \\
\bottomrule
\end{tabular}
\caption{Performance of Siamese Model on MOSEI and IEMOCAP} 
\label{siamese_accuracy}
\vspace{-3mm}
\end{table}

\section{Analysis and Ablation Studies} \label{sec:ablation_study}

\begin{table}[]
\renewcommand{\arraystretch}{0.7}
\setlength\tabcolsep{4.1pt}
\centering
\small
\begin{tabular}{ccccc}
\toprule
\multirow{2}{*}{Input Features} & \multicolumn{2}{c}{Classification} & \multicolumn{2}{c}{Emotion Shift} \\
\cmidrule{2-5}
                                & Accuracy         & F1              & Accuracy        & F1              \\
\midrule
G(L), O(A), F(V)                & 80.85            & 80.31           & 63.38           & 62.15           \\
B(L), O(A), O(V)                & 81.98            & 80.91           & 64.01           & 63.30           \\
B(L), O(A), O(V)                & \textbf{82.66}   & \textbf{83.07}  & \textbf{72.65}  & \textbf{67.32} \\
\bottomrule
\end{tabular}

\caption{Effect of different feature combinations for MOSEI. The classification columns are results on 2 class sentiment prediction task and emotion shift columns are results on 2 class emotion shifts classes. Here G(L): Glove, B(L): BERT, F(V): Facet, O(A): OpenSmile, O(V): OpenFace2.0}
\label{table:experiment-modelType}
\vspace{-3mm}
\end{table}
Due to a wide variety of components, it becomes vital to perform a detailed analysis of the architecture to understand the importance of various choices. 

\noindent\textbf{Feature and Design choices:}
For understanding the importance of features used for different modalities, we choose two different sets of features for text and visual modalities. In one setting, we use averaged GloVe embeddings \cite{glove} for text, OpenSmile features \cite{opensmile} for Audio and Facet features \cite{facet} for Video. Whereas in another setting, for text modality, we make use of a pre-trained  BERT \cite{bert} model's output layer. We calculate the average of the output layer to get a fixed-sized vector. For visual modality, we use the features provided by OpenFace2.0 \cite{openface2.0}, which are useful in performing facial analysis tasks such as facial landmark detection, head-pose tracking, and eye-gaze tracking. The results in Table \ref{table:experiment-modelType} (first two rows) highlight the advantage of features used in second setting.

\begin{table}[t]
\renewcommand{\arraystretch}{0.9}
\setlength\tabcolsep{5pt}
\centering
\small

\begin{tabular}{ccc} 
\toprule
{Emotion shift type} & Multilogue-Net & Our Model  \\ 
\midrule\
Positive - Negative                          & 69.78                   & \textbf{73.83}               \\
Negative - Positive                          & 59.49                   & \textbf{80.35}               \\
\bottomrule
\end{tabular}
\caption{Accuracy comparison with Multilogue-Net on MOSEI emotion shift utterances 
\label{table:emotion_shift}}
\vspace{-3mm}
\end{table}






\begin{table}[t]
\renewcommand{\arraystretch}{0.9}
\setlength\tabcolsep{5pt}
\centering
\small
\begin{tabular}{ccc}
\toprule
{Emotion shift type}  & 4-label  & 6-label  \\ \midrule
Positive - Negative              & 63.04             & 53.22             \\
Negative - Positive              & 69.77             & 70.68             \\ \bottomrule
\end{tabular}
\caption{Accuracy of the proposed model on IEMOCAP emotion shift utterances 
\label{table:emotion_shift_iemocap}}
\vspace{-6mm}
\end{table}

\noindent\textbf{Importance of Pretraining the Emotion Shift Component:}
To review the significance of the pretraining emotion shift component, we compare it with the two settings described above. We argue that jointly optimizing the emotion classification and emotion shift component from scratch might degrade the model classification performance. At the onset of training, the Siamese component does not provide a helpful signal to the classification component due to the random initialization of its weights, hampering the learning of the classification component. To prevent this, we pre-train the emotion shift component on the emotion shift labels separately before the joint training task, which helps provide the classification component with a better emotion shift signal at the start of training, making learning more accessible. The results in Table \ref{table:experiment-modelType} (third row) shows an increase of approximately  2\% in the F1 score when compared to the same features setting without pretraining (second row). Moreover, the Siamese network, when pre-trained, also achieves an F1 score of 67.32\%, the highest among all the experiments depicting the hindrance caused by joint training from scratch.

\noindent{\textbf{Performance over emotion shift utterances:}}
To verify the effectiveness of the emotion shift component we consider cases where an emotion shift has occurred between a target utterance $u_t$ and the prior utterance $u_{t-1}$ if there is a switch from positive emotion in $u_{t-1}$ to negative emotion in $u_t$, or vice versa (\S\ref{sec:methadology}). We evaluate our emotion classification performance on such utterances $u_t$ displaying an emotion shift.
Popular architectures like CMN, ICON, IANN and DialogueRNN perform poorly on the utterances with an emotion shift \cite{survey_paper}. In particular, in cases where the emotion of the target utterance differs from the previous utterance, DialogueRNN could only correctly predict 47.5\% instances, much lesser than the 69.2\% success rate that it achieves at the regions of no emotional shift. In Table \ref{table:emotion_shift} we compare our results with another multimodal ERC SOTA: Multilogue-Net. 
The results show a significant increase in accuracy for both positive to negative and negative to positive emotion shifts on the CMU-MOSEI dataset depicting the importance of the independent emotion shift component introduced in our architecture. Even though the Siamese network can predict the presence of emotion shift with an accuracy of about 72.65\% (Table \ref{table:experiment-modelType}), the signal received from it (in the form of reset and update gates of GRU) helps the emotion classification network to overcome the emotional inertia and predict the correct emotion. We also show the accuracy of our model on emotion shift utterances of the IEMOCAP dataset in Table \ref{table:emotion_shift_iemocap}. We could not calculate these numbers for CHFusion (SOTA on IEMOCAP) due unavailability of their code.

\noindent\textbf{Effect of Modeling Emotion Shift:}
To further verify the significance of modeling emotion shift as a separate component, we compare two variants of our best model - one with and the other without the Emotion shift component. Table \ref{table:experiment-emotionShift} shows a comparison on both the datasets. Across both the datasets, we observe a substantial increase in performance while using the emotion shift component. 







\begin{table}[t]
\renewcommand{\arraystretch}{0.9}
\setlength\tabcolsep{5pt}
\centering
\small
\resizebox{\columnwidth}{!}{
\begin{tabular}{cccccc} 
\toprule
\multicolumn{2}{c}{\multirow{2}{*}{Datasets}} & \multicolumn{2}{c}{Without} & \multicolumn{2}{c}{With}  \\ 
\cmidrule{3-6}
\multicolumn{2}{c}{}                          & A     & F1                  & A     & F1                \\ 
\midrule
\multicolumn{2}{c}{CMU-MOSEI}                 & 81.31 & 81.01               & \textbf{82.66} & \textbf{83.07}             \\
\multirow{2}{*}{IEMOCAP} & 4-label            & 77.53 & 77.51               & \textbf{78.47} & \textbf{78.46  }           \\
                         & 6-label            & 61.37 & 61.65               & \textbf{66.73 }& \textbf{66.86}             \\ 
\bottomrule
\end{tabular}
}
\caption{Performance with and without the emotion shift component (all numbers in \%)}
\label{table:experiment-emotionShift}
\vspace{-3mm}
\end{table}

\begin{table}[t]
\centering
\resizebox{\columnwidth}{!}{
\begin{tabular}{cccccccccc} 
\toprule
\multicolumn{2}{c}{\multirow{2}{*}{Datasets}} & \multicolumn{2}{c}{L+A} & \multicolumn{2}{c}{L+V} & \multicolumn{2}{c}{A+V} & \multicolumn{2}{c}{L+A+V}  \\ 
\cmidrule{3-10}
\multicolumn{2}{c}{}                          & A     & F1              & A     & F1              & A     & F1              & A     & F1                 \\ 
\midrule
\multicolumn{2}{c}{CMU-MOSEI}                 & 82.49 & 82.81           & 81.65 & 82.16           & 72.22 & 71.28           & \textbf{82.66} & \textbf{83.07  }            \\
\multirow{2}{*}{IEMOCAP} & 4-label            & \textbf{80.06} & \textbf{80.07 }          & 77.52 & 77.56           & 78.26 & 78.20           & 78.47 & 78.46              \\
                         & 6-label            & 64.26 & 64.48           & 63.34 & 63.40            & 58.90  & 58.84           & \textbf{66.73} & \textbf{66.86 }             \\
\bottomrule
\end{tabular}
}
\caption{Ablation study to observe the contribution of different modalities}
\label{table:experiments-modalities}
\vspace{-4mm}
\end{table}

\noindent\textbf{Contributions of the Modalities:}
To understand the importance of different modalities present in the datasets, we conduct experiments by choosing a combination of two out of the three modalities. As expected, models using all three modalities outperform models using only two modalities across most datasets (Table \ref{table:experiments-modalities}). On the IEMOCAP dataset with four classes, the text+audio model performs better than six classes. The text modality seems to be the most essential compared to other modalities highlighting the significance of context. 

\noindent\textbf{Using other modalities in the Emotion Shift Component:} 
To observe the effectiveness of modalities other than text on the Emotion Shift Component, we empirically analyze the effect of using all three modalities for training this component. We make use of the early fusion technique where modalities $l_t, a_t, v_t$ are concatenated ($l_t \oplus a_t \oplus v_t$ ) and then passed to the Siamese network. Observing the obtained results, we see that the use of the three modalities does not lead to an improvement (Table \ref{table:siamese_modality_ablation}). A possible reason for this might be the importance of context (captured in the text modality) in predicting emotion shifts. 

\begin{table}[t]
\renewcommand{\arraystretch}{0.9}
\setlength\tabcolsep{5pt}
\centering
\small

\begin{tabular}{ccc} 
\toprule
Modalities & Accuracy & F1     \\ 
\midrule
L                                                                              & \textbf{82.66}    & \textbf{83.07}  \\
L+A+V                                                                          & 82.20     & 82.78   \\
\bottomrule
\end{tabular}
\caption{Performance using other modalities in Siamese component (all numbers in \%)}
\label{table:siamese_modality_ablation}
\vspace{-4mm}
\end{table}



\noindent\textbf{Analyzing reset gate updates in GRU$_{arc}$:}
We also verify the practical significance of the emotion shift component qualitatively. We compare the GRU reset gate activations obtained from the emotion shift component and the reset gate activations learned by GRU without explicit emotion shift information. We randomly pick an instance from the CMU-MOSEI test set and analyze the GRU unit using it. In Figure \ref{fig:reset_gate_analysis}, we show these activations for the Video ID "m7SJs73SF8w" randomly selected from the test set. This dialogue has four utterances, and we see a shift from positive to negative emotion between utterances two and three and a shift from negative to positive emotion between utterances three and four. As seen in the left graph in Figure \ref{fig:reset_gate_analysis}, the emotion shift component learns to set a low reset gate value when there is an emotion shift (namely timestamps $t=3$ and $t=4$). This low reset gate value helps to weigh down the contribution of the previous emotion state for the predictions at the current timestamp. Comparing it to the case when we remove the emotion shift component (right graph in Figure \ref{fig:reset_gate_analysis}), the reset gate activations learned by the GRU do not follow the same trend, indicating that the previous emotion state will still significantly contribute to predictions at the current timestamp. Overall, the emotion shift component plays a vital role in effectively controlling information from the past.
\section{Discussion}

\begin{figure}[t]
\centering
\includegraphics[scale=0.42]{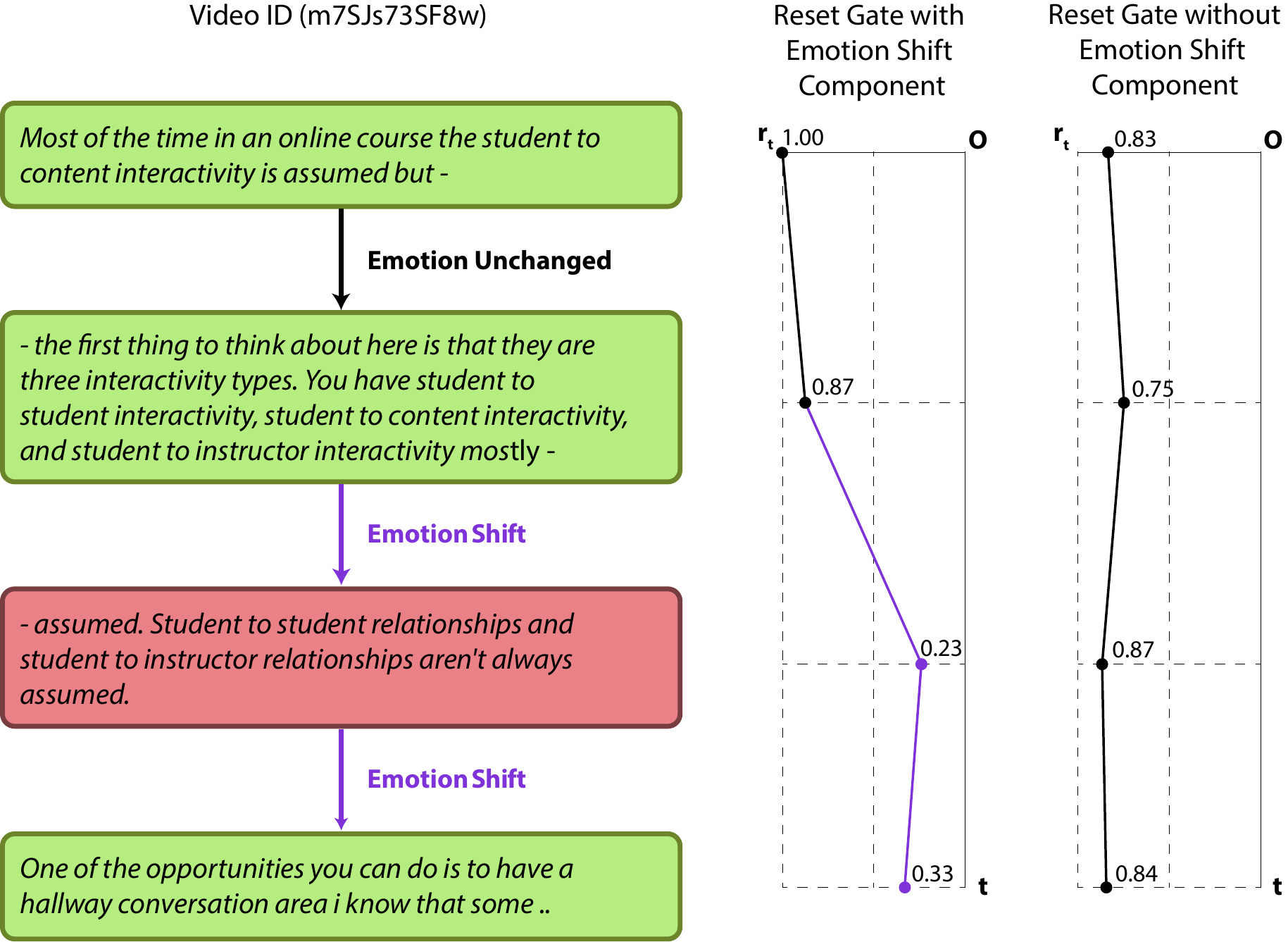}
\caption{Reset Gate activations on the dialogue `m7SJs73SF8w" from CMU-MOSEI test dataset}
\label{fig:reset_gate_analysis}
\vspace{-6mm}
\end{figure}

\noindent The presence of emotion shifts in human-to-human conversation is prominent in the conversational datasets. The existing works based on sequential modeling often suffer from these shifts, leading to poor performance for utterances with emotion shifts. In this work, we try to control the effect of previous utterances using an independent emotion shift module. As highlighted in Tables \ref{table:emotion_shift} and \ref{table:emotion_shift_iemocap}, the proposed architecture performs significantly better on emotion shift cases when compared to Multilogue-Net (20\% improvement in negative-positive and 4\% improvement on positive-negative shifts). The novel design of the emotion shift-based gating mechanism in the GRU unit helps boost the prediction performance for utterances with emotion shifts. As noticed in Fig. \ref{fig:reset_gate_analysis}, the reset and update gates provide a significant signal when there is an emotional shift in conversation.

\noindent \textbf{Modularity:} The modular design and idea of the proposed emotion shift component can further be used to improve any emotion prediction systems that have poor performance in emotion shift cases. Moreover, the designed emotion shift component works considering only the textual modality, making it applicable to both multimodal as well as unimodal systems. 

\noindent \textbf{Application to Real-Time Systems:} A notable limitation of all the existing Emotion Recognition state-of-the-art systems often comes from the incapability of their implementations for real-time use cases as they require the entire context to be given in the form of multiple utterances to the model. For future approaches where the models will target the real-time setting, the proposed emotion shift component can be handy as it only uses two consecutive utterances to predict the emotion shift.

\section{Conclusion and Future Directions}
\noindent In this paper, we proposed a deep learning based model for  multimodal emotion recognition in conversations. We proposed a new emotion shift component (modeled using the Siamese net) that captures the emotional arc in a conversation and steers the main emotion recognition model. We performed a battery of experiments on two main emotion recognition datasets. Results and analysis show the importance of the emotion shift component. Currently, the emotion shift component uses only the text modality for predicting the shift and we plan to explore more sophisticated ways of using information from multiple modalities. 

\section{Acknowledgements}
We would like to thank reviewers for their insightful comments. This research is supported by SERB India (Science and Engineering Board) Research Grant number SRG/2021/000768.

\bibliography{references}
\bibliographystyle{acl-natbib}

\clearpage
\newpage
\appendix
\section*{Appendix}

\section{Evaluation Metrics} \label{appendix:metrics}
Consider $\bc{y_n}_{n=1}^N$ as the true labels and $\bc{\hat{y}}_{n=1}^N$ as the predicted labels for the $N$ datapoints. Note that $y_n, \hat{y}_n \in \bc{1, 2, \ldots, K}$ where $K$ is the number of classes.

The \textbf{Accuracy score} for predictions if given by:\\
\begin{align*}
    \text{Accuracy} = \frac{\sum_{n=1}^N\mathbb{I}\bs{y_n = \hat{y}_n}}{N}
\end{align*}
We use the \texttt{accuracy\_score} method of Python based Scikit-learn library \cite{sklearn} for its evaluation.
The \textbf{F1 score} for a class $k$ is given by:\\
\begin{align*}
    \text{F1}_k = \frac{2\times\text{precision}_k\times\text{recall}_k}{\text{precision}_k+\text{recall}_k}
\end{align*}\\
where, precision$_k$ is the precision for class $k$ and recall$_k$ is the recall for class $k$. These are calculated using:
\begin{align*}
    \text{precision}_k = \frac{\underset{\hat{y}_n = k}{\sum}\mathbb{I}\bs{y_n = \hat{y}_n}}{\underset{\hat{y}_n = k}{\sum}1} \\
    \text{recall}_k = \frac{\underset{{y}_n = k}{\sum}\mathbb{I}\bs{y_n = \hat{y}_n}}{\underset{{y}_n = k}{\sum}1}
\end{align*}

Finally, the \textbf{weighted F1 score} is defined as
\begin{align*}
    \text{weighted F1} = {\sum_{k=1}^K{f_k\times \text{F1}_k}}
\end{align*}
where $f_k$ is the relative frequency of class $k$ 

We use the \texttt{F1\_score} method of Scikit-learn library for its evaluation.





\section{Experiment Reproducibility} \label{appendix:hyper}
\subsection{Input and hidden states dimensions}
The Input modality dimensions for the different datasets we experimented are as follows:

\noindent\textbf{CMU-MOSEI}
        \begin{itemize}
            \item Text (BERT): 768
            \item Audio (OpenSmile): 384
            \item Video (OpenFace2.0): 711
        \end{itemize}
\noindent\textbf{IEMOCAP:}
        \begin{itemize}
        \item Text (BERT): 768
        \item Audio (OpenSmile): 100
        \item Video (3D CNN): 512
        \end{itemize}
    
The dimension of the hidden states and GRU states are as follows:
\begin{itemize}
    \item Siamese hidden state ($\cH_{t}$): 300
    \item Party state for each modality $s^{q_t,m}_t$: 150 
    \item Context State for each modality $c^m_t$: 150
    \item Emotion State for each modality $e^m_t$: 100
\end{itemize}
All other weights and parameters are such that the equations given in $\S\ref{sec:methadology}$ hold.

There are a total of 5578803 parameters in the model.

\subsection{Training the main model}
All experiments are implemented using the PyTorch library \cite{pytorch}. All weights are initialized randomly using PyTorch's default methods, and we use the Adam optimizer \cite{adam} for training these weights.

The following hyper-parameters are used for the optimizer:
\begin{itemize}
    \item Learning rate (lr) : 0.0001
    \item Weight decay (weight\_decay): 0.0001
    \item $\beta_1, \beta_2$ (betas): (0.9, 0.999)
\end{itemize}
Here, the names in parenthesis denote the arguments corresponding to the hyper-parameters in the Adam Optimizer object of the PyTorch library.

We use a batch size of 128 for training across all experiments. The number of epochs for which the model was trained varies across datasets. These are listed as follows:
\begin{itemize}
    \item CMU-MOSEI - 50 epochs 
    \item IEMOCAP - 500 epochs
\end{itemize}

Training time per epoch was approximately 2.5 minutes for CMU-MOSEI and 15 seconds for IEMOCAP.

The model is evaluated at every epoch on the validation set (constructed using an 80:20 random split of the training data). The model giving the best weighted average F1 score across all classes is checkpointed. All the randomizations in the training procedure are reproducible using a seed value of 42 for libraries NumPy and PyTorch.

\subsection{Training of the emotion shift component}
This section provides the hyper-parameters for the pre-training procedure of the emotion shift component described in \ref{subsec:proposed_model}. We use a batch size of 8, and the model is pre-trained for five epochs. The model is checkpointed against the best F1 score.

\subsection{Hyperparameter Tuning}
Hyperparameters like the size of siamese hidden state ($\cH_t$), size of context/party/hidden states ($s^{q_t,m}_t,c^m_t,e^m_t$) are tuned manually. The best weighted average F1 score over the validation set across all epochs was used as the criterion to select the best hyperparameter configuration.

To tune the hyperparameters used in the optimizer (learning rate, weight decay, $\beta_1$, $\beta_2$), we started with the default values used in the PyTorch library. These values are:
\begin{itemize}
    \item learning rate: 0.001
    \item weight decay: 0
    \item $\beta_1$: 0.9
    \item $\beta_2$: 0.999
    
\end{itemize}
On manual tuning, we found that decreasing the learning rate to $0.0001$ and increasing the weight decay to $0.0001$ helped in better convergence and superior validation performance. Changing the values of $\beta_1$ and $\beta_2$ did not lead to any improvement. So these were kept the same as the default values.
\subsection{Machine Specification}
All experiments were performed on a server using Intel i7-5820K CPU @ 3.30GHz, Nvidia GeForce GTX TITAN X GPU, and CUDA 11.




\end{document}